\documentclass[article,12pt]{elsarticle}
\usepackage{natbib}
\usepackage{graphicx}
\usepackage{amsmath,amssymb,latexsym}

\newcommand{\bp}{\mbox{\boldmath $p$}}

\newcommand{\be}{\begin{eqnarray}}
\newcommand{\ee}{\end{eqnarray}}
\newcommand{\bee}{\begin{eqnarray*}}
\newcommand{\eee}{\end{eqnarray*}}

\newcommand{\matrixb}{\left[ \begin{array}}
\newcommand{\matrixe}{\end{array} \right]}

\journal{ }

\begin{document}

\begin{frontmatter}

% Title.
% ------
\title{Active Search for Nearest Neighbors}
%
% Single address.
% ---------------
%\author{Daniel Fisher}

%\address{dan.fisher@live.com}

\author{Hayoung Um and Heeyoul Choi}

\address{Handong Global University \\
558 Handong-ro, Pohang, Korea 37554 \\
{\tt \{ejklektov, heeyoul\}@gmail.com}}

\begin{abstract}
In pattern recognition or machine learning, it is a very fundamental task to find nearest neighbors of a given point.  All the methods for the task work basically by comparing the given point to all the points in the data set. That is why the computational cost increases with the number of data points. However, the human visual system seems to work in a different way. When the human visual system tries to find the neighbors of one point on a map, it directly focuses on the area around the point and actively searches the neighbors by looking or zooming in and out around the point. In this paper, we propose an innovative search method for nearest neighbors, which seems very similar to how human visual system works on the task. 
\end{abstract}


\end{frontmatter}

\section{Introduction}

In pattern recognition or machine learning, it is a very fundamental task to find nearest neighbors of a given point \cite{Duda2001book}. 
As stated by manifold hypothesis, a real world data set lies on a concentrated low-dimensional nonlinear subspace in a high-dimensional space \cite{Rifai2011nips}, and the subspace can be characterised by topology which is roughly defined as a locally Euclidean space. With data samples, any locally Euclidean space at a point is implmented by nearest neighbors of the point.

Many machine learning algorithms like Isomap and locally linear embedding are based on nearest neighbors \cite{Tenenbaum2000science,Roweis2000science,HChoi2007pr}. Once the nearest neighbors are selected, the algorithms find a better representation preserving the local properties in a lower dimensional space. Also, $k$ Nearest Neighbors ($k$NN) is a popular classification algorithm that relys on nearest neighbors. 

As big data has become more popular, it requires more computation time to find neighbors out of a huge number of data points. 
To find neighbors quickly, many algorithms have been proposed, including KD-tree \cite{Bentley1975acm} and locality sensitive hashing (LSH) \cite{Indyk1998tc}, which provide approximated solutions. 
All the methods need a computational time based on the number of data points, $N$, even though the time could be $\log(N)$.  
They work basically by comparing the given point to all the points in the data set. That is why the computational cost increases as does $N$. Actually, the methods have to check all the points that are represented as vectors on the Cartesian coordinates. 

However, the human visual system works in a different way, not based on the vectors on Cartesian coordinates. When the human visual system tries to find the neighbors of one point on a map, it directly focuses on the area around the point and actively searches the neighbors by looking or zooming in and out around the point. That is, the human visual system takes the data set as an image on the retina space with selective attention. The computational time for the human visual system for the task is almost independent of $N$. 

In this paper, we propose an innovative algorithm to actively search $k$ nearest neighbors. The proposed algorithm transforms the vectors on the Cartesian coordinates into an image and then search the neighbors on the image. 
The method does not check all the data points but only the points that are around the given point, which makes  the computational cost independent of the number of data points. 

%This approach is conceptually related to embedded cognition or extended mind \cite{Noe2009book}, in a sense that the data is placed out of the agent.  

\section{Active Search Algorithm}

\begin{figure}[h]
\centering{\includegraphics[width=1.3in]{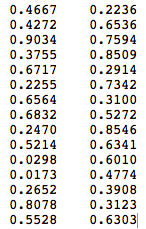}\hspace{0.55in}
\includegraphics[width=2.25in]{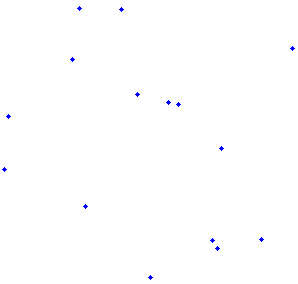}}
\caption{(Left) 15 data points as 2 dimensional vectors on Cartesian coordinates, and (Right) an image of the points.}
\label{fig:transform}
\end{figure}

The problem can be formulated as follows. Given a data set with $N$ points (as vectors), an search algorithm should find $k$ nearest neighbors of a new point $\bp$. 
To follow how human visual system works on the task, we propose an active search algorithm which interpretes the data points as an image, as shown in Fig.\ \ref{fig:transform}. 
We can see that the image is easier than the vectors to understand properties of the data set, especially when it comes to data distribution.

The proposed algorithm starts from the location of the point $\bp$ on the image, and actively searches the neighbor points by zooming in and out around the location. 
That is, the algorithm transforms $\bp$ onto the same image as the $N$ data points, and determines its location on the image. Then it checks all the image pixels within a circle with a radius $r$, at the location.  

%Draw a picture of pixels. %%%%%%%

\begin{figure}[h]
\centering{\includegraphics[width=3.25in]{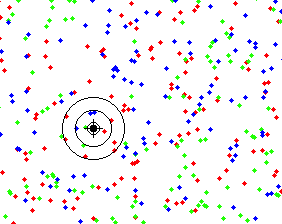}}
\caption{Active search on an image for the neighbors of a new point, presented as the plus (`+') mark. The color of the points represents the same class.}
\label{fig:search}
\end{figure}

The radius is initialized as $r_0$ and increases (or decreases) if the number of points within the current circle, $n_t$, is smaller (or greater) than $k$. Since $n_t$ is proportional to the area of the circle, basically the radius is modified iteratively by 
\be
r_{t+1} = \mbox{round} \left(r_t * \sqrt{\frac{k}{n_t}}\right),
\label{eq:radius}
\ee
where $\mbox{round}( \cdot )$ is the rounding operation to make the radius working on the pixel unit. 
If $n_t$ is equal to $k$, the algorithm returns the points within the circle as the $k$ nearest neighbors. The update of the radius is illustrated in Fig.\ \ref{fig:search}. %Other strategies could be applied to change the radius. 
To show that the proposed algorithm can work with classification tasks, class information in a data set is represented by different colors on the image as in Fig.\ \ref{fig:search}. 

It is important to find a proper image resolution. The sparsity of data points on the image affects the accuracy of our method compared to the original $k$NN method. There is a trade-off between the computation time and the accuracy. If the data points are transformed onto a low resolution image, some points might overlap with another ones, which would make our method inaccurate to count the number of points. If the resolution increases, the algorithm requires a bigger memory size and has to check more pixels which needs a more computational cost. 

When it comes to classification, this problem can be alleviated by using as many images as the number of classes, each of which images represents the data points from one class and each pixel keeps the number of data points on it. But still it is inevitable to have approximated solutions if the resolution of the image is not high enough to represent all the points as individual dots on the image, although in real world problems, exactly overlapping points from different classes are rare cases.

\section{Numerical Experiments}
We compared our active search algorithm to the original $k$NN algorithm, using randomly generated 2 dimensional data points as in Fig.\ \ref{fig:search}. 
Given the number of classes is 3, the two algorithms classify 100 new points based on 11 nearest neighbors. The original $k$NN algorithm is considered as the ground truth for the accuracy of the proposed method. 
For the active search algorithm, the data points were transformed into a 3000$\times$3000 square image, and the initial radius, $r_0$ was set to 100 pixels. 

\begin{figure}[h]
\centering{\includegraphics[width=4.25in]{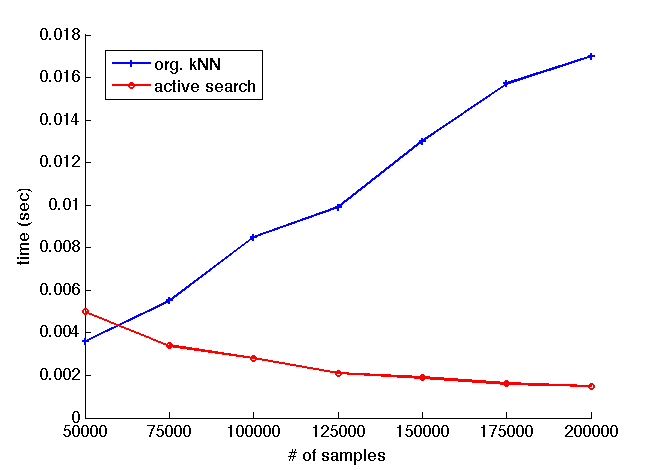}}
\caption{For original $k$NN (blue crosses) and the proposed active search (red circles), elapsed times (sec) on 2.7 GHz Intel Core i5. The image resolution is fixed to $3000\times3000$ for all cases.}
\label{fig:times}
\end{figure}

First, we checked the time elapsed on the algorithms with different number of data points, which is shown in Fig.\ \ref{fig:times}. 
We can see that the time for the original $k$NN increases linearly with $N$. Even though the most efficient algorithm could take only $\log(N)$ for the best case, it also increases with $N$.   

However, the proposed method even decreases while $N$ increases. Actually our method is supposed to be independent of $N$. The reason it decreases is that the sparser the data points are on the image, the longer time the method takes to find exact number of points increasing the radius, because the initial radious was fixed to 100, which seems too small. 

Most of the computational cost comes from checking all the inner pixels of the current circle on the image to find neighbors based on the Euclidean distance. When the L1 distance is taken, the computational cost could be extremely cheap, while the result would be more roughly approximated than the Euclidean distance, compared to the original $k$NN. 

%Note that since transforming the training data points onto the image is one time job as a preprocessing, it is not counted in the time elapsed. 

Currently, compared to the original $k$NN, the accuracy of the proposed method on the randomly generated 2 dimensional data points is up to 98$\%$. Note that this $k$NN on the randomaly generated points is the worst case for classification in a sense that there is no class structure in the data set. 
This approach can be applied to higher dimensional data, though it will require a much bigger memory (or disk) space. 
%Once transforming the data into an image, image tools like Google image search API could be helpful in handling such a high dimensional image \cite{GoogleAPI}. 

\section{Conclusion}
We proposed an active search algorithm for $k$ nearest neighbors, which transforms the data points onto an image and searches the neighbors on the image, as the visual system in the human brain does. The computational time of the proposed algorithm is almost independent of the number of data points. This method can be considered as an alternative approach for searching nearest neighbors, especially when computational time is crucial. 
%It is a future work to extend the approach to high dimensional data sets. 

%************* references ***********************

%\section*{References}
\bibliographystyle{elsarticle-num}
\bibliography{hchoi}
\end{document}